\documentclass[letterpaper, 10 pt, conference]{ieeeconf}  % Comment this line out if you need a4paper

\IEEEoverridecommandlockouts                              % This command is only needed if
\usepackage{graphics} % for pdf, bitmapped graphics files
\usepackage{epsfig} % for postscript graphics files
\usepackage{mathptmx} % assumes new font selection scheme installed
\usepackage{times} % assumes new font selection scheme installed
\usepackage{amsmath} % assumes amsmath package installed
\usepackage{amssymb}  % assumes amsmath package installed
\usepackage{graphicx}
\usepackage{mathtools}
\usepackage{xcolor}
\usepackage{multirow}
\usepackage{float}
\usepackage{colortbl}
\usepackage{makecell}

\title{\LARGE \bf
3D Object Aided Self-Supervised Monocular Depth Estimation
}

\author{Songlin Wei, Guodong Chen*, Wenzheng Chi*, Zhenhua Wang and Lining Sun % <-this % stops a space %
\thanks{*Corresponding author}% <-this % stops a space
\thanks{Research was partially supported by the National Key R\&D Program
of China grant \#2019YFB1310201.}%
\thanks{The authors are with the Robotics and Microsystems Center,
School of Mechanical and Electric Engineering, Soochow University, Suzhou 215021, China \{{\it slwei, wzchi, chenguodong, wangzhenhua, lining.sun}\}@suda.edu.cn}%
}

\begin{document}

\maketitle
\thispagestyle{empty}
\pagestyle{empty}

%%%%%%%%%%%%%%%%%%%%%%%%%%%%%%%%%%%%%%%%%%%%%%%%%%%%%%%%%%%%%%%%%%%%%%%%%%%%%%%%
\begin{abstract}
Monocular depth estimation has been actively studied in fields such as robot vision, autonomous driving, and 3D scene understanding. Given a sequence of color images, unsupervised learning methods based on the framework of Structure-From-Motion (SfM) simultaneously predict depth and camera relative pose. However, dynamically moving objects in the scene violate the static world assumption, resulting in inaccurate depths of dynamic objects. In this work, we propose a new method to address such dynamic object movements through monocular 3D object detection. Specifically, we first detect 3D objects in the images and build the per-pixel correspondence of the dynamic pixels with the detected object pose while leaving the static pixels corresponding to the rigid background to be modeled with camera motion. In this way, the depth of every pixel can be learned via a meaningful geometry model. Besides, objects are detected as cuboids with absolute scale, which is used to eliminate the scale ambiguity problem inherent in monocular vision. Experiments on the KITTI depth dataset show that our method achieves State-of-The-Art performance for depth estimation. Furthermore, joint training of depth, camera motion and object pose also improves monocular 3D object detection performance. To the best of our knowledge, this is the first work that allows a monocular 3D object detection network to be fine-tuned in a self-supervised manner.
\end{abstract}

%%%%%%%%%%%%%%%%%%%%%%%%%%%%%%%%%%%%%%%%%%%%%%%%%%%%%%%%%%%%%%%%%%%%%%%%%%%%%%%%
\section{Introduction}
Monocular depth estimation is a challenging task for 3D scene understanding. Research on monocular depth estimation has grown rapidly due to its relatively low cost, ease of deployment, and freedom from complex calibration. Among them, unsupervised learning based methods reformulate the problem of depth estimation as an image reconstruction problem by using the photometric consistency between consecutive image frames as a supervisory signal. Current self-supervised learning based methods for depth estimation can even achieve comparable accuracy to supervised learning methods, which makes unsupervised depth estimation very attractive because it avoids the laborious collection of large amounts of data in real-world environments. The unsupervised neural network typically consists of two sub-networks for learning the depth map and the relative pose between consecutive frames, respectively. By using the estimated depth of the target image and the motion of the camera to transform nearby frames, a reconstructed view is obtained to compare with the target view. Then the photometric error between the reconstructed view and the target view is used to train the whole network. However, when there are dynamically moving objects in the scene that violate the static world assumption, the accuracy of predicted depth is affected.
% should metion old methods not considering dynamic objects?

\begin{figure} [t]
    \centering
    \includegraphics[width=\linewidth]{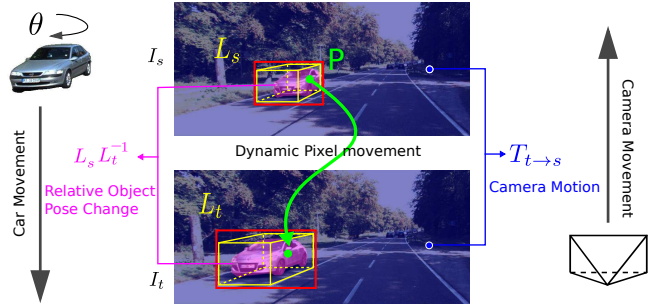}
    \caption{The dynamic pixel movement between target view $I_t$ and source view $I_s$ is explicitly modeled with the detected 3D object poses $L_t$ and $L_s$, while the static pixel is modeled with $T_{t{\rightarrow}s}$. In this way, the depth of every pixel can be learned via a meaningful geometric model. Detail description is given in Section \ref{ssec_handledynamic}.}
    \label{fig_idea}
\end{figure}

In order to address this problem, some researchers such as \cite{zhou2017unsupervised} \cite{mahjourian2018unsupervised} try to eliminate the impact of dynamic objects by masking out the pixels which are unexplainable by the camera motion. However, the pixels belonging to the dynamic objects are not modeled. Another work \cite{yang2018every} employs FlowNet \cite{dosovitskiy2015flownet} to model the movement of dynamic pixels directly and revise the loss function accordingly. However, the accuracy of depth estimation is limited with the performance of the unsupervised FlowNet.
% Nevertheless, the network still suffers from the occlusion and disocclusion caused by dynamic objects.
Different from these methods, in this work, we propose to exploit 3D object detection to model the dynamic moving objects explicitly, as illustrated in Fig. \ref{fig_idea}. By modeling the movement of pixels on the dynamic object, the depth of the dynamic pixels will no longer be coupled with the camera motion $T$ to formulate the appearance loss but rather the object pose $L$. Then the depth of the dynamic pixels can be learned via a meaningful appearance loss.

Moreover, scale ambiguity is inherent in monocular vision, i.e., the depth and translation of the relative camera pose between two frames can only be recovered up to an unknown scale. Guizilini et al. \cite{guizilini20203d} proposed to use the velocity of the camera to constrain the scale. The velocity can be obtained through the IMU unit. However, velocity information has accumulated drift over time, for example, when the camera is moving at a constant speed. In this work, we propose to exploit the absolute scale of detected objects to constrain the translation of camera movement so as to recover the true scale.
\begin{figure*}
    \centering
    \includegraphics[width=0.95\textwidth]{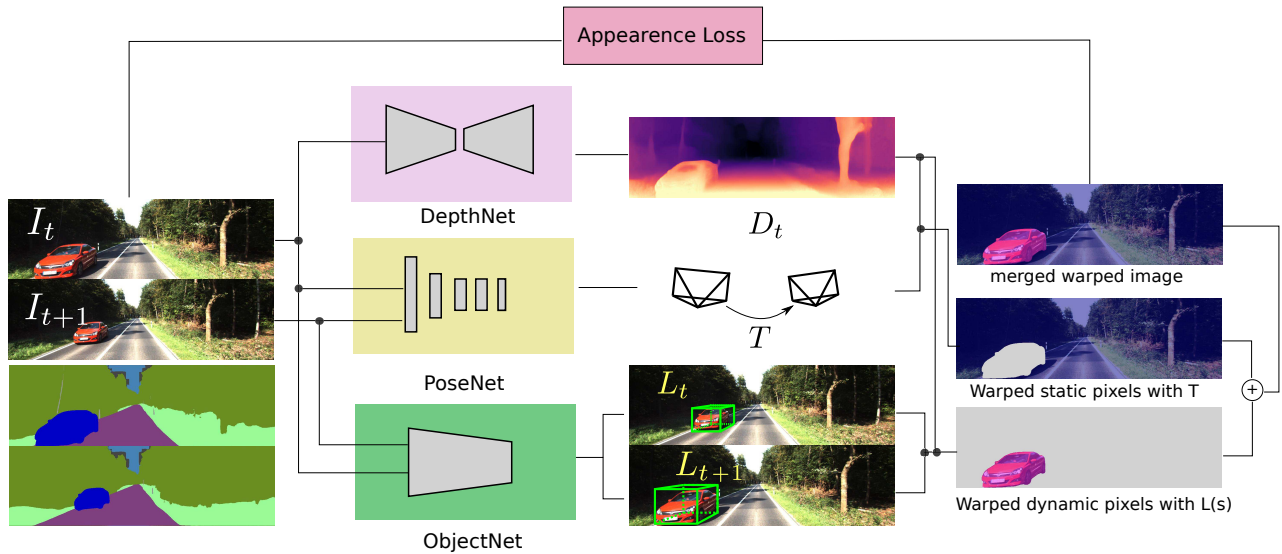}
    \caption{Overview of the network. Image $I_t$ is passed into DepthNet to predict depth map $D_t$. Both target view $I_t$ and source views $I_s$ are passed into PoseNet to predict the camera relative pose $T$, and the two views are also passed into ObjectNet to detect objects in each view. The reconstructed image is warped with $D_t$, $T$ and $L$(s) for static background and dynamic objects, respectively. Finally, the appearance loss of the reconstructed image and target image is calculated. The whole network can then be trained by minimizing the appearance loss. Especially, the ObjectNet can be trained along with DepthNet and PoseNet in an unsupervised manner.}
    \label{fig_networks}
\end{figure*}

After introducing 3D object detection into the monocular depth estimation problem, the photometric loss between the warped and target images can be back-propagated to the 3D object detection network. Therefore, 3D object detection networks can be fine-tuned in a self-supervised manner. This has been shown in experiments to benefit both object detection and depth estimation.

Our main contributions are summarized as follows:

\begin{itemize}
    \item We novelly introduce monocular 3D object detection into the self-supervised depth estimation problem to address the dynamic object movements. After we detect 3D objects in each view, the per-pixel correspondence can be built either by relative object pose change or camera motion. In this way, the depth of every pixel can be learned via a meaningful geometric model.
    \item A novel scale ambiguity elimination method is proposed with the absolute scale of the detected 3D objects.
    \item The 3D object detection network is fine-tuned in a self-supervised manner along with depth and pose estimation networks.
\end{itemize}

\section{Related Work}
Simultaneously estimating geometric structure and camera relative pose has long been well studied in Structure-from-Motion. Whilst Deep learning is recently adopted to deal with more challenging scenarios where traditional feature exaction and matching pipeline fails. The most appealing aspect of deep learning methods is that it learns depth knowledge as a prior and can estimate the dense map given a single image without predicting the camera pose during test time.

\subsubsection{Self-supervised Monocular Depth Estimation}
Instead of directly regressing depth values using convolutional neural networks as supervised methods do, self-supervised methods redefine depth prediction as an image reconstruction problem. Garg et al. \cite{garg2016unsupervised} proposed an autoencoder structured network to predict the depth map of an image. Then, the fixed displacement between the two views of the binocular camera is used to warp one of the views, and the photometric error of the reconstructed view and with the opposite view is used to train the network. However, this approach produces "texture-copy" artifacts and depth discontinuities. To address this issue, Godard et al. \cite{godard2017unsupervised} proposed to predict both left-to-right and right-to-left disparities and further enforce the spatial coherence between them, thereby improving performance. Zhou et al. \cite{zhou2017unsupervised} extend the paradigm to monocular vision. The authors propose to use temporal image sequences to simultaneously predict single-view depth and multi-view camera pose via separate DepthNet and PoseNet, which quickly became the de facto standard for monocular depth estimation. Mahjourian et al. \cite{mahjourian2018unsupervised} and Liu et al. \cite{liu2019unsupervised} employ geometric models from Structure-from-motion (SfM) to estimate camera ego-motion instead of PoseNet. Guizilini et al. \cite{guizilini20203d} proposes a novel 3D packing and unpacking layer to better preserve the structural details of the input color image during feature shrinkage and expansion, however at the cost of higher computation. Our work adopts canonical PoseNet and DepthNet structures in \cite{zhou2017unsupervised}, and introduces additional object detection networks.

\subsubsection{Dynamic Objects Handling}
The unsupervised learning based depth estimation method works well in static scenes. However, the presence of dynamic moving objects brings extra challenges, as it breaks the static world assumption. Liu et al. \cite{liu2019unsupervised} proposed a unified framework for jointly learning optical flow and depth to segment rigid regions. Then the photometric and geometric loss are built on rigid region to avoid the influence of dynamic regions. Klingner et al. \cite{klingner2020self} applied cross-task training for both semantic segmentation and depth estimation. The feature map learned from semantic segmentation is proved to be beneficial for depth estimation. Besides, the pixels which are labeled with dynamic class objects are also masked out. Other works try to model the dynamic objects explicitly. Yang et al. \cite{yang2018every} propose to use an optical flow network to build correspondences of pixels between frames including dynamic pixels. The authors model the dynamic objects with 3D scene flow. Casser et al. \cite{casser2019unsupervised} propose to take advantage of semantic labels to mask out the objects, which are then fed into a separate similar network with PoseNet to predict the object motion separately. With the predicted motion of the objects, the geometric reprojection model can again be applied to dynamic objects.

Different from these works, we use a monocular 3D object detector \cite{brazil2019m3d} to detect objects in each frame and model the dynamic pixels directly with the detected object poses, as illustrated in Fig. \ref{fig_idea}. We also take advantage of the absolute scale of the detected 3D objects to propose a novel scale loss to constrain the PoseNet to predict the right scale.

\section{Method}
Depth estimation is the problem of estimating the dense depth map $D_t$ given a target view image $I_t$ at time $t$, as illustrated in Fig. \ref{fig_networks}, where $D_t$ has the same resolution as $I_t$ with each pixel representing the point distance from the camera. Self-supervised depth estimation tackle this problem with a novel view-synthesis method. The temporally preceding and succeeding frames denoted as $\{I_{t-1}, I_{t+1}\}$ are feed together with $I_t$ into the network for predicting their relative transformation matrices with frame $I_t$, denoted as $T_{t{\rightarrow}s} \in SE(3)$, $s \in \{\textnormal{-}1, 1\}$. Suppose the camera intrinsic matrix is $K \in \mathbb{R}^{3\times3}$, and the homogeneous coordinate of a pixel in the target view is denoted as $\bar{p_t}$. Following the work of \cite{zhou2017unsupervised} \cite{babu2018undemon}, we can obtain the projected homogeneous pixel coordinates in the source image $\bar{p_s}$ as:
\begin{equation} \label{eqn0}
    \bar{p_s} = K \; T_{t{\rightarrow}s} \; D(p_t) \; K^{-1} \; \bar{p_t}
\end{equation}
However, this model only works for the static points in the world. Next, to address this problem, we describe our method of modeling the dynamic points with the detected 3D object poses.

\begin{figure} [t]
    \centering
    \includegraphics[width=0.6\linewidth]{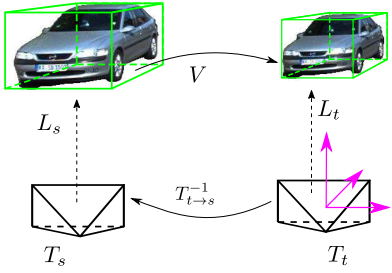}
    \caption{Coordinate System: The target frame $T_t$ is set as the reference frame. The solid curves illustrate the camera and object movements with arrows indicating the transformation direction. The dotted lines with arrows illustrate the single view object detections. The relationship of the transformation matrices of different poses is given in Equation (\ref{eqn1}).}
    \label{fig_coords}
\end{figure}

\subsection{Modeling Dynamic Moving Points}
\label{ssec_handledynamic}
To ease illustrating the geometric relationships of camera frames and object frames, the world coordinate system is built at the target image frame, as illustrated in Fig. \ref{fig_coords}. The camera pose at time $s$ is denoted as $T_{s}$. Suppose an object $O$ is observed both in the source and target frame with poses $L_s$ and $L_t$ respectively. Further suppose it moves from time $s$ to $t$ with transformation matrix $V$. Then we have
\begin{equation}
    L_t = T_{t{\rightarrow}s}^{-1} \; L_{s} \;  V . \label{eqn1}
\end{equation}
For a physical point $P=[x,y,z,1]^T$ in the object frame, the position of $P$ relative to the object does not change assuming the object is rigid. Then the 3D coordinate of point $P$ at different times can be obtained by
\begin{equation}
    P_t = L_t \; P \label{eqn2}
\end{equation}
\begin{equation}
    P_{s} = L_t \; V^{-1} \; P \label{eqn3}
\end{equation}
The coordinates are different when the object is moving, that is, $V$ is not the identity matrix. Then the homogeneous coordinate of the corresponding pixel in view $I_s$ can be acquired through projection:
\begin{equation}
    \bar{p}_{s} = K \; T_{t{\rightarrow}s} \; P_{s}, \label{eqn5}
\end{equation}
Notice that there is an implicit conversion from homogeneous coordinates to non-homogeneous coordinates before multiplying to K. After substituting $V$ from (\refeq{eqn1}) and $P$ from (\refeq{eqn2}) into (\refeq{eqn3}) and further substituting $P_s$ from (\refeq{eqn3}) into (\refeq{eqn5}), we have
\begin{equation}
    \bar{p}_{s} = K \; L_{s} \; L_t^{-1} \; D(p_t) \; K ^{-1} \; \bar{p_t} \label{eqn7}
\end{equation}
where $D(p_t) \; K ^{-1} \; \bar{p_t} =P_t$ is the back projected 3D coordinates for pixel $p_t$. Comparing (\refeq{eqn7}) with (\refeq{eqn0}), we can find that the projected pixel of dynamic object points in source frames dependent \emph{only} on object poses from 3D object detector. It does not dependent on camera ego-motion $T$ nor object motion $V$. Indeed, when the object is not moving, then $T_{t{\rightarrow}s}$ and $L_s L_t^{-1}$ are equal according to (\ref{eqn1}). Therefore, the per-pixel correspondence between $\bar{p}_{s}$ and $\bar{p_t}$ which corresponding to dynamic objects is established through object pose change $L_s L_t^{-1}$.

\subsection{Self-Supervised Learning for Depth Estimation}
After the rigid background pixels and the dynamic object pixels are projected into the source frames, the synthesized target view $\tilde{I}_{s{\rightarrow}t}$ can be acquired through sampling from each source view $I_s$:
\begin{equation}
    \tilde{I}_{s{\rightarrow}t} = M^{(0)} \; I_s \langle p_s^{(0)} \rangle + \sum_{i=1} M^{(i)} \; I_s \langle p_s^{(i)} \rangle ,
\end{equation}
where the superscript $(0)$ denotes the pixel from the rigid background and $(i)$ denotes the pixel from the object $i$, and $M^{(0)}$ is the binary mask of the rigid background, $M^{(i)}$ is the binary mask of the $i$-th object. All the masks are acquired in the source image, therefore not overlapping with each other. $\langle\rangle$ is a differentiable bilinear sampling operator. $p_s$ can be computed with (\refeq{eqn0}) if the pixel is static, and (\refeq{eqn7}) if it is dynamic. Now the photometric error between the composite view and the target view is formulated as
\begin{equation}
    \mathcal{L}_{ap} = \min_{s} \mathit{pe}(I_t, \tilde{I}_{s{\rightarrow}t})
\end{equation}
Herein, $\mathit{pe}$ is the appearance similarity function and the minimum reprojection loss of all source views for each point is applied. The function $pe$ is defined as:
\begin{equation}
    \mathit{pe}(I_t, \tilde{I}_{s{\rightarrow}t}) = \frac{\alpha}{2} (1 - \mathit{SSIM}(I_t, \tilde{I}_{s{\rightarrow}t})) + (1-\alpha) \Vert I_t - \tilde{I}_{s{\rightarrow}t}\Vert_1,
\end{equation}
which considers both $\mathit{SSIM}$ \cite{wang2004image} and $L1$ differences. $\alpha$ is a balancing constant. In projection geometry, the same color image corresponds to numerous different real scenes in the physical world. To make the depth estimation problem solvable, an edge-aware smoothness regularization term \cite{godard2017unsupervised} is adopted here:
\begin{equation}
    \mathcal{L}_{sm} = \vert \partial_x d_t^* \vert exp(-\partial_x I_t) + \vert \partial_y d_t^* \vert exp(-\partial_y I_t)
\end{equation}
where $d_t^*$ is mean value normalized depth of $D_t$. We follow Godard et al. \cite{godard2019digging} to predict the depth map at multiple image scales. Finally, the combined loss function is formulated as:
\begin{equation}
    \mathcal{L}_{photo} = \lambda \mathcal{L}_{sm} + \sum_l \mathcal{L}_{ap}^l \label{eqn11},
\end{equation}
where $\lambda$ is a weighting constant. The superscript $l$ denotes the appearance loss is calculated at image scale $l$. Then, each subnetwork can be trained by minimizing the final loss $\mathcal{L}_{photo}$ by gradient descent algorithms.

\begin{figure} [t]
    \centering
    \includegraphics[width=\linewidth]{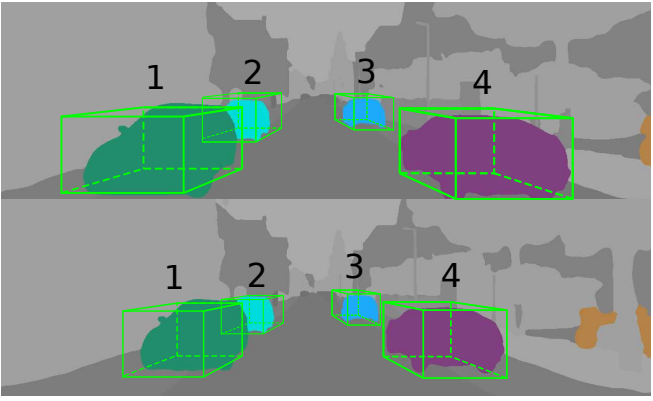}
    \caption{An example of data association for objects detected in both target view (Up) and source view (Bottom). Each detected object is shown as a cuboid indexed with numbers. Objects with the same numbers are associated.}
    \label{fig_association}
\end{figure}

\begin{figure*}
    \centering
    \includegraphics[width=\textwidth]{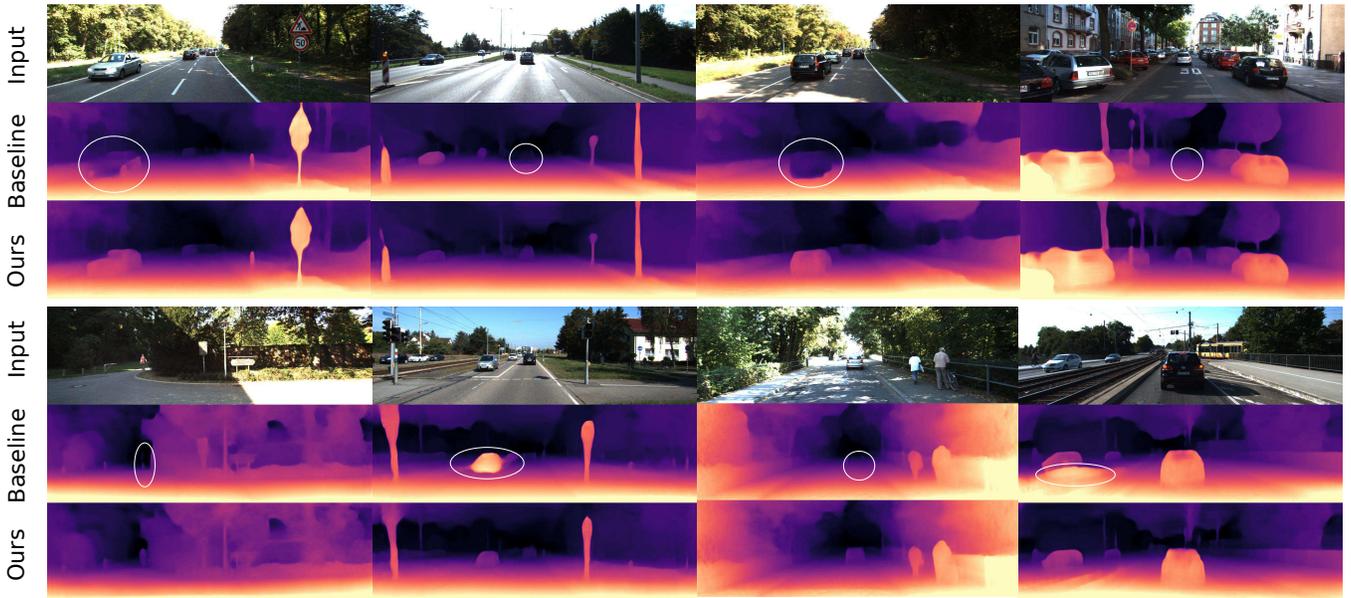}
    \caption{Qualitative comparison of our method with Godard et al. \cite{godard2019digging} as the baseline. Every 3 rows show the input image, baseline depth, and ours. Despite the fact that the baseline method can handle moving objects e.g. cars most of the time. There are still some frames observed failing to predict the depth of moving objects while our method works reliably as marked in the circles.}
    \label{fig_qualitydynamic}
\end{figure*}

\subsection{Monocular 3D Object Detection}
Now we discuss how to acquire the object pose in the source and target views. We use a 3D region proposal network similar to \cite{brazil2019m3d} and follow the convention of autonomous driving literature to use 3D cuboids to represent objects in monocular images. Each cuboid $C$ is comprised of 2D bounding box $B_{2d}=[x, y, w, h]_{2d}$, 3D bounding box $B_{3d}=[w, h, l]_{3d}$, center location $t_c=[x, y, z]_c$ and rotation angle $\theta$ with respect to the camera frame, as illustrated in Fig. \ref{fig_idea}. The object pose $L$ relative to the camera frame is
\begin{equation}
    L = \begin{bmatrix}
        \cos \theta & 0 & \sin \theta & x_c \\
        0 & 1 & 0 & y_c \\
        \sin \theta & 0 & \cos \theta & z_c \\
        0 & 0 & 0 & 1
        \end{bmatrix}.
\end{equation}
Before the detected object pose $L$ can be used to model the dynamic point, the objects detected in every single source and target view must be associated. The problem is also known as data association. Following Wei et al. \cite{9700775}, we consider both the object cuboid shape, location, and 2D box for the association. The score $s$ for each potential association is calculated as
\begin{equation}
    s(C^i, C^j) = IoU_{2D}(C^i, C^j) + D(C^i, C^j) + \alpha \; S(C^i, C^j),
\end{equation}
where $\alpha = 0.5 $ for balancing the similarity for object distance and shapes. $IoU_{2D}$ is the Intersection over Union (IoU) of the two 2D boxes, defined as:
\begin{equation}
    D(C^i, C^j ) = exp(-|t_{c}^i-t_{c}^j|),
\end{equation}
which considers the object centroid distance. While $S$ is defined as
\begin{equation}
    S(C^i, C^j) = \sum_{s \in w,h,l} exp(-|B^i_{3d}[s]-B^j_{3d}[s]|),
\end{equation}
which considers the shape differences of each dimension. As shown in Fig. \ref{fig_association}, each object that appeared in the target view is associated with the object wih the highest score in the source view. Then, the projection model proposed in (\refeq{eqn7}) can be applied to all the associated objects.

To further improve the precision of the projection model, we take advantage of panoptic segmentation to distinguish whether every pixel in the 2D box is on or off the object. As shown in Fig. \ref{fig_idea}, some pixels in the 2D bounding box are actually from the background. To remedy this problem, we use the off-the-shelf tool Panoptic-DeepLab \cite{panoptic_deeplab_2020} to perform panoptic segmentation for each view. All the training images are preprocessed and are fed together with the color images into our network.

\subsection{Scale Recovery with Object Poses}
The detected object poses and sizes have true scale because the object detection network is trained with the real size and distance of objects in similar scenes. As a result, the estimated depths of dynamic points that minimize the reprojection appearance loss are consistent with the true scale. For the static points, we design a scale loss to constrain the translation of the camera ego-motion instead of using the additional camera velocity information as proposed in \cite{guizilini20203d}. Specifically, we check the reprojection error of objects with camera ego-motion $T_{s,t}$ against the relative object pose change $L_sL_t^{\textnormal{-}1}$ . If the two errors are close enough, the object is then determined as static since the reprojection error can also be explained with the camera ego-motion. Then we use the translation of static objects to constrain the translation of the camera. This can be justified by Equation (\refeq{eqn1}), as $L_sL_t^{\textnormal{-}1}$ equals to $T_{t{\rightarrow}s}$ for static objects with $V$ set to identity matrix. Therefore, we propose the scale loss as follows:
\begin{equation}
    \mathcal{L}_{scale} = \vert tran(T_{t{\rightarrow}s}) - \frac{1}{N} \sum_{i} tran(L_s^{(i)}L_t^{(i)\textnormal{-}1}) \vert
\end{equation}
where $N$ is the number of detected static objects and $tran$ is the function to extract the translation part from a transformation matrix. The translation part of object pose changes is averaged over all static objects. Finally, the total loss function is
\begin{equation}
    \mathcal{L}_{final} = \mathcal{L}_{photo} + \beta \mathcal{L}_{scale}
\end{equation}
where $\mathcal{L}_{photo}$ is from Equation (\refeq{eqn11}) and $\beta$ is a weighting constant set to $0.05$. The scale loss part is effective to recover the true scale of the predictions without decreasing the network performance.

\section{Experiments}

\subsection{Network Structure and Implementation Details}
Our network is comprised of three subnetworks, DepthNet, PoseNet, and ObjectNet, as illustrated in Fig. \ref{fig_networks}. The three subnetworks are connected by the appearance loss between the reconstructed view and the target view. Specifically, the reconstructed view is segmented into two parts: static background and dynamic objects. The dynamic part is obtained by warping the source with $D_t$ together with the detected object pose $L$ by ObjectNet. While the static part is obtained by warping the source view $D_t$ together with $T$ predicted by PoseNet. The DepthNet and PoseNet are similar with \cite{zhou2017unsupervised}. The DepthNet is an auto-encoder like structure for single view depth estimation. It accepts a single color image and outputs inverse depth map $Disp_t$ with the same resolution. The inverse depth is converted to depth map $D_t$ by $1/(a * Disp_t + b)$, with specifically chosen parameters $a$ and $b$ to constrain the depth to be in the valid range $[0.1,80]$ for the KITTI dataset \cite{Geiger2012CVPR}. The ResNet18 \cite{he2016deep} with pre-trained weights on ImageNet \cite{5206848} is used as the encoder part for feature extraction. The decoder part has 4 layers of building blocks each consisting of convolution and upsampling. The PoseNet accepts two temporally connected image sequence and outputs the 6 DoF relative camera pose $T_{t{\rightarrow}s}$. The 3D object detection network is similar to \cite{brazil2019m3d} which adopts DenseNet121 \cite{huang2017densely} as backbone feature extractor and followed by a Region Proposal Network (RPN) layers that use 3D anchor boxes to perform monocular 3D object detection.

Our proposed network is implemented with PyTorch \cite{paszke2017automatic} framework. The panoptic segmentation of color images are precomputed with \cite{panoptic_deeplab_2020}. The ObjectNet is also pretrained on KITTI \textit{object} dataset \cite{Geiger2012CVPR}. Regarding data augmentation for DepthNet and PoseNet, we use common random horizontal flips and color augmentation with random brightness, contrast, saturation, and hue dithering, ranging from $\pm 0.2$, $\pm 0.2$, $\pm 0.2$ and \cite{godard2019digging} proposed $\pm 0.1$. ObjectNet has random flipping disabled, but detected object poses are flipped when necessary. The network is trained for 15 epochs with a learning rate of 0.0001 and drops to 0.00001 for the last 5 epochs. Training takes about 48 hours on an Nvidia RTX 3090 24G graphic card. ObjectNet is frozen for the first 15 epochs to make training more stable. In the last 5 epochs, ObjectNet is fine-tuned together with DepthNet and PoseNet, excluding feature extraction layers to reduce computational cost.

\subsection{Depth Evaluation on KITTI Dataset}
There are total 61 scenes in KITTI raw dataset \cite{Geiger2012CVPR} consisting of city, residential and road. Following Eigen et al. \cite{eigen2014depth} and Zhou et al. \cite{zhou2017unsupervised}, a total number of 39810 training images and 4424 validation images from both the left and right camera with static frames removed are used for training. The test set contains 697 images from different scenes from the training set. We compare our work with other networks using similar depth encoder layers. A qualitative comparison of depth estimation with baseline Monodepth2 \cite{godard2019digging} is given in Fig. \ref{fig_qualitydynamic}. We can find that our network can better predict the depth of dynamic objects as marked in the figure. In some highly dynamic scenes, our network can predict reliable depth for moving objects while the baseline network fails in some frames. We also quantitatively evaluate the depth estimation results with standard metrics as defined in \cite{eigen2014depth}. A detailed comparison with other state-of-the-art work is shown in Table \ref{tab_depthmetric}. Our method achieves improved performance without scaling the depth with median value, which is commonly adopted by other works. Following the convention on KITTI depth evaluation, only certain area $A=\{(x,y) | x \in (0.03594771, 0.96405229), y \in (0.40810811, 0.99189189) \}$ of meaningful points in the depth map is used when calculating all the metrics, where the pixel coordinates are normalized to $[0,1]$ according to the depth map width and height.

\begin{table*}[h]
    \caption{Quantitative Results for depth evaluation on KITTI using Eigen Split. All the works to be compared are monocular methods which are trained with temporal images only. We follow the evaluation method provided by \cite{zhou2017unsupervised} and the depth is capped at 80 meters. All the results in the table are using the same evaluation method without further online refinements.}
    \label{tab_depthmetric}
    \begin{center}
    \begin{tabular}{|c||c c c c||c c c|}
    \hline
      & \multicolumn{4}{c||}{Error Metrics}  &  \multicolumn{3}{c|}{Accuracy Metrics} \\
    Method & Abs Rel & Sq Rel & RMSE & RMSE log & $\delta < 1.25$ & $\delta < 1.25^2$ & $\delta < 1.25^3$ \\
    \hline
    Zhou et al. \cite{zhou2017unsupervised} (SfMLearner) & 0.183 & 1.595 & 6.709 & 0.270 & 0.734 & 0.902 & 0.959 \\
    \hline
    Yang et al. \cite{yangunsupervised2018} & 0.182 & 1.481 & 6.501 & 0.267 & 0.725 & 0.906 & 0.963 \\
    \hline
    Mahjourian et al. \cite{mahjourian2018unsupervised} & 0.163 & 1.240 & 6.220 & 0.250 & 0.762 & 0.916 & 0.968\\
    \hline
    Zhan et al. \cite{zhan2018unsupervised} & 0.135 & 1.132 & 5.585 & 0.229 & 0.820 & 0.933 & 0.971 \\
    \hline
    Casser et al . \cite{casser2019unsupervised} & 0.141 & 1.025 & 5.28 & 0.215 & 0.816 & 0.945 & 0.979 \\
    \hline
    Godard et al. \cite{godard2019digging} (Monodepth2) & 0.115  &   0.903  &   4.863  &   0.193  &   0.877  &   0.959  &   0.981 \\
    \hline
    Klingner et al. \cite{klingner2020self} (SGDepth full) & \cellcolor{gray!25}0.113 & 0.835 & \cellcolor{gray!25}4.693 & 0.191 & 0.879\cellcolor{gray!25} & 0.961 & 0.981 \\
    \hline
    Guizilini et al. \cite{guizilini2019semantically} & 0.117 & 0.854 & 4.714 & 0.191 & 0.873 & \cellcolor{gray!25}0.963 & 0.981 \\
    \hline
    Ours & 0.115  &   \cellcolor{gray!25}0.831  &   4.734  & \cellcolor{gray!25}0.190\cellcolor{gray!25}  &   0.877  &   0.960  &   \cellcolor{gray!25}0.983 \\
    \hline
    \end{tabular}
    \end{center}
\end{table*}

\begin{table*}[h]
    \caption{Ablation study of our model on KITTI \cite{geiger2013vision} dataset using Eigen split \cite{eigen2014depth} . The baseline is adopted from \cite{godard2019digging}. All the metrics are obtained through the evaluation method \cite{zhou2017unsupervised}. Notice that the predicted depths are capped at 80 meters.}
    \label{tab_ablation}
    \begin{center}
    \begin{tabular}{|c|c|c|c||c c c c||c c c|}
    \hline
      & & & & \multicolumn{4}{c||}{Error Metrics}  &  \multicolumn{3}{c|}{Accuracy Metrics} \\
    Method & \makecell{Object \\ Detection} & \makecell{Scale \\ Loss} & \makecell{Joint \\ Training} & Abs Rel & Sq Rel & RMSE & RMSE log & $\delta < 1.25$ & $\delta < 1.25^2$ & $\delta < 1.25^3$ \\
    \hline
    Ours baseline & & & & \cellcolor{gray!25}0.115  &   0.903  &   4.851  &   0.193  &   0.877  &   0.959  &   0.981 \\
    \hline
    Ours (w/o scale loss) & \checkmark & & & 0.126  &   0.879  &   4.799  &   0.198  &   0.844  &   0.957  &   0.982 \\
    \hline
    Ours (w/o joint fine-tuning) & \checkmark & \checkmark & & 0.116  &   0.844  &   4.758  &   0.190  &   0.871  &   \cellcolor{gray!25}0.960  &   \cellcolor{gray!25}0.983 \\
    \hline
    Ours (full) & \checkmark & \checkmark & \checkmark & \cellcolor{gray!25}0.115  &   \cellcolor{gray!25}0.831  &   \cellcolor{gray!25}4.734  & \cellcolor{gray!25}0.190\cellcolor{gray!25}  &   \cellcolor{gray!25}0.877  &   \cellcolor{gray!25}0.960  &   \cellcolor{gray!25}0.983 \\
    \hline
    \end{tabular}
    \end{center}
\end{table*}    

\begin{table*}[h]
    \caption{Different metrics on val1 data split for Cars before and after fine-tuning.}
    \label{tab_finetuning}
    \begin{center}
    \begin{tabular}{|c||c c c||c c c||c c c|}
        \hline
        \multirow{2}{*}{Tasks} & \multicolumn{3}{c||}{$AP_{BEV}$ (IoU $\ge$ 0.7)} & \multicolumn{3}{c||}{$AP_{3D}$ (IoU $\ge$ 0.7)} & \multicolumn{3}{c|}{2D Detection}  \\
        \cline{2-10}
        & Easy & Mod & Hard & Easy & Mod & Hard & Easy & Mod & Hard \\
        \hline
        Before & 0.2416 & \cellcolor{gray!25}0.1937 & 0.1589 & 0.1808 & 0.1447 & 0.1235 & 0.8865 & 0.8260 & 0.6719\\
        \hline
        After & \cellcolor{gray!25}0.2422 & 0.1844 & \cellcolor{gray!25}0.1642 & \cellcolor{gray!25}0.1846 & \cellcolor{gray!25}0.1511 & \cellcolor{gray!25}0.1292 & \cellcolor{gray!25}0.8971 & \cellcolor{gray!25}0.8357 & \cellcolor{gray!25}0.6758 \\
        \hline
    \end{tabular}
    \end{center}
\end{table*}

\subsection{Scale Recovery and Pose Evaluation}
\label{ssec_scalerecovery}
\begin{figure}[h]
    \includegraphics[width=\linewidth] {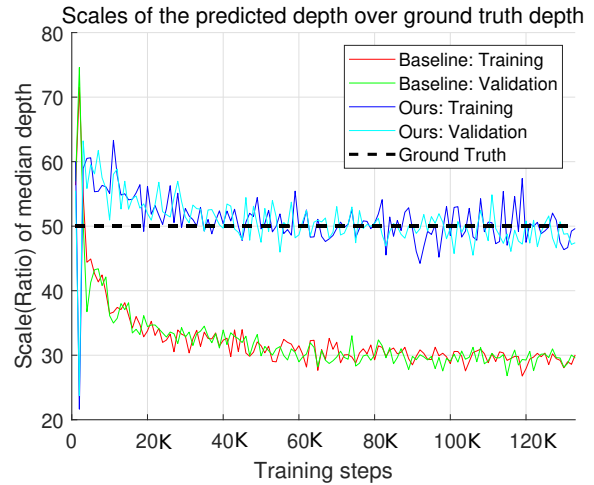}
    \caption{The $scale$ of the ground truth depth map and the predicted depth map. The predicted scale of our network converges to the true scale $50$ marked with red dotted line after about 3 epochs while the baseline network converges around $30$.}
    \label{fig_trainvalscale}
\end{figure}

To verify the proposed network predicting the right scale in the depth map, we record the scale at each training step. The scale is defined as $scale=med(D_t^*)/med(D_t)$, where $D_t^*$ and $D_t$ are the ground truth depth map and the predicted depth map, respectively and $med$ is the median value operator. The $med$ is taken over the entire batch of inputs with batch size as $6$. To ease the numerical convergence of the network, we scale the detected objects with $0.02$, which constrains the right scale to be $50$. As shown in Fig. \ref{fig_trainvalscale}, the predicted scale of our network is stabilized at 50 while the baseline network converges to the wrong scale.

We also performed an ablation study of how scale loss affects the depth estimation as shown in Table. \ref{tab_ablation}. We first train the network without scale loss and manually set the object scale to a fixed number. Not surprisingly, the results are even a little worse than baseline because the scales of the camera motion and the detected objects are inconsistent. The scale of camera motion (translation) is from PoseNet and the scale of object pose is from a separate ObjectNet. Without scale loss, the depths of rigid background and dynamic objects which are indirectly determined by the scales of camera motion and objects respectively are with different scales, hence leading to a worse result. With scale loss which enforces the predicted scales are consistent, our model is performed as expected.

\subsection{Fine-Tuning for 3D Object Detection}
\begin{figure} [t]
    \includegraphics[width=\linewidth]{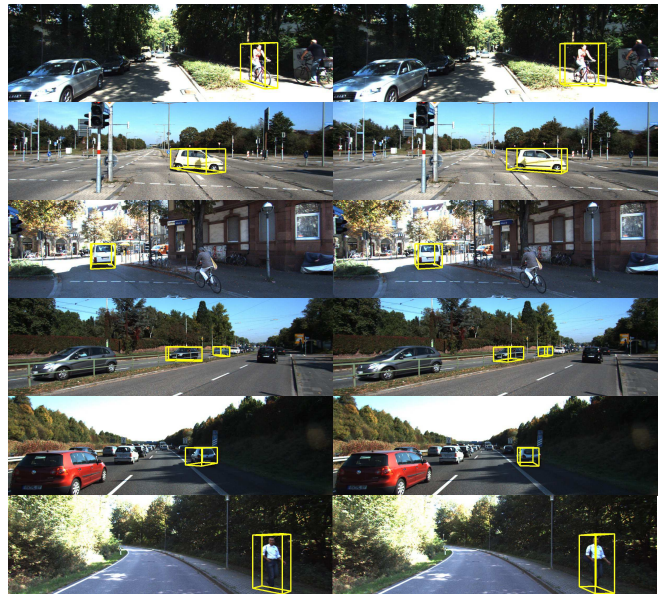}
    \caption{We observed orientation angles refinements over some detected objects. Left column shows the cuboid detected before fine-tuning. Right columns show the refinements after fine-tuning.}
    \label{fig_finetuning}
\end{figure}

The 3D object detection network is fine-tuned in a self-supervised manner with depth estimation. After $5$ epochs of fine-tuning together with DepthNet and PoseNet, we observed improvements in object orientation angles, as shown in Fig. \ref{fig_finetuning}. Some objects which are partially occluded are detected more consistently on orientations. We also reevaluate the fine-tuned ObjectNet on the original KITTI \textit{3D object} dataset. The evaluation is conducted on the val1 \cite{brazil2019m3d} data splits. We compared the Average Precision (AP) \cite{simonelli2019disentangling} on 3 core tasks including Bird Eye View (BEV), 3D Object detection and 2D detection under 3 different settings: Easy, Moderate and Hard \cite{simonelli2019disentangling}\cite{Geiger2012CVPR}. The AP is calculated with IoU $\ge$ 0.7. All the metrics are improved after fine-tuning as shown in Table \ref{tab_finetuning}. This shows the effectiveness of our proposed self-supervised training of monocular 3D object detection.

\subsection{Ablation study}
A detailed ablation study of our model is given in Table. \ref{tab_ablation}. We adopt Monodepth2 \cite{godard2019digging} (monocular variation) as the baseline and gradually apply object detection, scale loss, and joint training technique to the baseline. We observed improvements when object detection is combined together with scale loss as discussed in Sec. \ref{ssec_scalerecovery}. Furthermore, the joint training of depth prediction and object detection improves the results even more as proposed.

\section{Conclusion}
We have presented a novel method for handling dynamically moving objects in monocular depth estimation. By modeling dynamic points on objects with detected object poses, every pixel in the scene can be explained by camera motion or object pose changes. The benefits of introducing object detection for depth estimation are threefold. First, explicit modeling of dynamic points is shown to have better depth estimation performance. Second, static object poses can be used to regulate the ego-motion of the camera. The result shows that our network not only achieves state-of-the-art monocular depth estimation accuracy but also produces the correct scale. Finally, the object detection network is also fine-tuned in a self-supervised manner, which is proven to benefit the original object detection network.

\bibliographystyle{IEEEtran}
\bibliography{IEEEtranBST/IEEEabrv,IEEEtranBST/mybibfile}
\end{document}